\pdfoutput=1

\documentclass[11pt]{article}

\usepackage[final]{neurips_2023}

\usepackage{latexsym}
\usepackage{CJKutf8}
\usepackage[utf8]{inputenc} 
\usepackage[T1]{fontenc}    
\usepackage{booktabs}       
\usepackage{amsfonts}       
\usepackage{nicefrac}       
\usepackage{microtype}      
\usepackage{xcolor}         
\usepackage{times}
\definecolor{linkColor}{rgb}{0.18,0.39,0.62}
\usepackage[colorlinks = true,
           linkcolor = linkColor,
           urlcolor  = linkColor,
           citecolor = linkColor,
           anchorcolor = linkColor]{hyperref}
\usepackage{soul}
\usepackage{latexsym}
\usepackage{amsmath}
\usepackage{makecell}
\usepackage{graphicx} 
\usepackage{multirow}
\usepackage{multicol}
\usepackage{algorithm}    
\usepackage{algpseudocode}
\usepackage{amssymb}
\usepackage{ucs}
\usepackage{enumitem}
\usepackage{color}
\usepackage{subcaption}
\usepackage{soul}
\usepackage{fontawesome}

\title{Extensible Prompts for Language Models on \\ Zero-shot Language Style Customization}

\author{
\textbf{Tao Ge ~~ Jing Hu ~~ Li Dong ~~ Shaoguang Mao ~~ Yan Xia ~~ Xun Wang} \\
\textbf{Si-Qing Chen ~~~~~ Furu Wei} \\
Microsoft \\
{ \texttt{\{tage,v-hjing,lidong1,shamao,yanxia,xunwang\}@microsoft.com }} \\
  \texttt{\{sqchen,fuwei\}@microsoft.com}
}

\newcommand{\ctext}[3][RGB]{%
  \begingroup
  \definecolor{hlcolor}{#1}{#2}\sethlcolor{hlcolor}%
  \hl{#3}%
  \endgroup
}

\begin{document}
\maketitle
\vspace{-0.6cm}
\begin{abstract}
We propose eXtensible Prompt (X-Prompt) for prompting a large language model (LLM) beyond natural language (NL). X-Prompt instructs an LLM with not only NL but also an extensible vocabulary of imaginary words. Registering new imaginary words allows us to instruct the LLM to comprehend concepts that are difficult to describe with NL words, thereby making a prompt more descriptive. Also, these imaginary words are designed to be out-of-distribution (OOD) robust so that they can be (re)used like NL words in various prompts, distinguishing X-Prompt from soft prompt that is for fitting in-distribution data. We propose context-augmented learning (CAL) to learn imaginary words for general usability, enabling them to work properly in OOD (unseen) prompts.
We experiment X-Prompt for zero-shot language style customization as a case study.
The promising results of X-Prompt demonstrate its potential to facilitate advanced interaction beyond the natural language interface, bridging the communication gap between humans and LLMs.
\end{abstract}
\vspace{-0.5cm}

\begin{table}[!h]
\centering
\small
\caption{X-Prompt introduces an extensible vocabulary of imaginary words to represent what NL words hardly describe. For example, the imaginary words \ctext[RGB]{255,255,0}{$\widetilde{w_{\textrm{satya}}}$} and \ctext[RGB]{97,189,252}{$\widetilde{w_{\textrm{sheldon}}}$} are learned from \textit{Satya Nadella}'s tweets and \textit{Sheldon Cooper}'s lines to denote their language style respectively. They can be used like NL words in various contexts to prompt an LLM to generate specified content in Satya's and Sheldon's styles, even if the specified content (e.g., criticize/praise the C++ language) has never been talked about by them before, achieving zero-shot style customization.}\label{tab:x-prompt}\vspace{0.1cm}
\begin{tabular}{c|p{4cm}|p{7cm}}
\toprule
\bf Prompt Type & \textbf{Prompt (shortened\protect\footnotemark)}                                 & \textbf{Text Generated by Prompting a Frozen LLM}                                                                                                                                                                                 \\ \midrule
NL & \textcolor{blue}{Criticize} the C++ language:      & \textcolor{blue}{ \textit{The C++ language is a mess. It is a mess because it is a language that is designed to be used by people who are not programmers ...}}                                                                                                      \\ \midrule
\multirow{2}{*}{X-Prompt} & \textcolor{blue}{Criticize} the C++ language in \ctext[RGB]{255,255,0}{$\widetilde{w_{\textrm{satya}}}$} style: & \textcolor{blue}{\textit{C++ is a great language, but it's not the right tool for every job. We're working on a new language for the next generation of Microsoft products ...}}   \\ \cline{2-3}
 & \textcolor{blue}{Criticize} the C++ language in \ctext[RGB]{97,189,252}{$\widetilde{w_{\textrm{Sheldon}}}$} style: &\textcolor{blue}{ \textit{Well, I’m not going to waste my time with C++, or any C-derived programming language. They’re all too hard. I’m sticking with my Python ...}}   \\ 
\midrule \midrule
 NL & \textcolor{red}{Praise} the C++ language:               & \textit{\color{red} C++ is a language that is easy to learn, easy to use, and easy to program. It is a language that is easy to ...}       \\ \midrule
 \multirow{2}{*}{X-Prompt} & \textcolor{red}{Praise} the C++ language in \ctext[RGB]{255,255,0}{$\widetilde{w_{\textrm{satya}}}$} style: & \textcolor{red}{\textit{C++ is a language that has been around for decades, and it's still the most widely used programming language in the world ...}} \\ \cline{2-3}
 & \textcolor{red}{Praise} the C++ language in \ctext[RGB]{97,189,252}{$\widetilde{w_{\textrm{sheldon}}}$} style:    & \textcolor{red}{\textit{Yeah, that's good. It's actually very close to the way I think about programming.}}  \\ 
 \bottomrule     
\end{tabular}
\end{table}

\footnotetext{Due to space limit, prompt texts in Table are shortened. Original prompt texts are presented in Appendix \ref{appsec:shorten}.}


\section{Introduction}\label{sec:intro}

Recent work~\citep{brown2020language} has observed language models (LMs) tend to be increasingly capable of in-context learning as their model size grows. The emergent capability~\citep{wei2022emergent} allows instructing a large LM at run time using a descriptive natural language (NL) prompt to solve a specified task with out-of-distribution (OOD) robustness~\citep{liu2022sample}.

Nonetheless, it is not always easy to come up with a descriptive prompt, especially for tasks involving fine-grain specifications that are beyond words. For example, it is hard to elaborate a person's language style using NL to prompt an LM to write in his/her language, unless it is well-known (e.g., \textit{William Shakespeare} style).

To provide access to delivering more descriptive prompts, we propose eXtensible Prompt (X-Prompt), inspired by Textual Inversion \citep{gal2022image}. Compared with NL prompts, X-Prompt additionally introduces an extensible vocabulary of imaginary words that are learned to represent what NL words hardly describe. 
For example, an imaginary word\footnote{In this paper, we use $\widetilde{w}$ to denote an imaginary word, as opposed to $w$ denoting a natural language word.} $\widetilde{w}_u$ representing a specific person $u$'s style can be combined with various prompt contexts to instruct the LM to generate specified content in $u$'s language, as shown in Table \ref{tab:x-prompt}. 

In contrast to soft prompt~\citep{qin2021learning} that is for fitting in-distribution (ID) data and thus is likely to fail in OOD prompting scenarios~\citep{vu2022spot,su2021transferability,lester2022reducing}, imaginary words in X-Prompt are designed to be OOD robust and well generalized so that they can be used like NL words for various (even unseen) prompt purposes, as illustrated in Table \ref{tab:intro_id_ood}.

\begin{table}[h]
\centering
\vspace{-0.3cm}
\caption{A comparison between soft prompt and X-Prompt in both ID and OOD prompt scenarios during inference (\textcolor{gray}{gray text} is the generated text by the model given the prefix). In contrast to soft prompt (i.e., the soft token \ctext[RGB]{0,220,252}{[SOFT]}) that works well in ID but performs poorly in OOD prompt scenarios, X-Prompt has significantly better OOD robustness, whose imaginary word (i.e., \ctext[RGB]{0,209,0}{$\widetilde{w_{\textrm{Trump}}}$}) can be used like an NL word in various contexts for different prompt purposes.}\vspace{0.1cm}\label{tab:intro_id_ood}
\small
\scalebox{0.89}{
\begin{tabular}{p{1.5cm}|p{5.5cm}|p{7cm}}
\toprule
\textbf{}                            & \multicolumn{1}{c|}{\textbf{Soft prompt}}                                                                                               & \multicolumn{1}{c}{\textbf{X-Prompt}}                           \\ \midrule
\multicolumn{1}{c|}{\textbf{Train}}                    & \ctext[RGB]{0,220,252}{{[}SOFT{]}} $\to$ \{Trump's text\}                                                                                            &  Write in \ctext[RGB]{0,209,0}{$\widetilde{w_{\textrm{Trump}}}$} style: $\to$ \{Trump's text\}                    \\ \midrule
  & \ctext[RGB]{0,220,252}{{[}SOFT{]}} \textit{Make \textcolor{gray}{America Great Again!!!}}   \faThumbsOUp                                                                                      & Write in \ctext[RGB]{0,209,0}{$\widetilde{w_{\textrm{Trump}}}$} style: \textit{Great job \textcolor{gray}{by @elonmusk.}}  \faThumbsOUp    \\ \cline{2-3}

\multicolumn{1}{c|}{\textbf{ID Prompt}} & \ctext[RGB]{0,220,252}{{[}SOFT{]}} \textit{\textcolor{gray}{Republicans should have gotten rid of Section 230 in the Defense Bill}}    \faThumbsOUp                                            & Write in \ctext[RGB]{0,209,0}{$\widetilde{w_{\textrm{Trump}}}$} style: \textit{Fantastic\textcolor{gray}{!! Thank you so much for your support!}}     \faThumbsOUp               \\ \midrule

    & \ctext[RGB]{0,220,252}{{[}SOFT{]}} Concern about cloning technology: \textit{\textcolor{gray}{Big voter fraud information coming out concerning Georgia. Stay tuned!}}  \faThumbsDown                                   & Concern about the cloning technology in \ctext[RGB]{0,209,0}{$\widetilde{w_{\textrm{Trump}}}$} style: \textit{\textcolor{gray}{The United States is a nation of laws. We do not clone people. We do not clone animals.}} \faThumbsOUp   \\ \cline{2-3}

\multicolumn{1}{c|}{\textbf{OOD Prompt}} & \ctext[RGB]{0,220,252}{{[}SOFT{]}} Criticize the C++ language: \textit{\textcolor{gray}{All the Radical Left Democrats are trying to do with the Post Office hearings is blame the Republicans for the FRAUD that will occur ...}} \faThumbsDown & Criticize the C++ language in \ctext[RGB]{0,209,0}{$\widetilde{w_{\textrm{Trump}}}$} style: \textit{\textcolor{gray}{C++ is a very difficult language to learn and understand. It is very complicated and full of bad habits, which will make it much more difficult to fix problems...}}\faThumbsOUp            \\ \bottomrule
\end{tabular}
}\vspace{-0.2cm}
\end{table}

To ensure the general usability of imaginary words, we propose context-augmented learning (CAL). It guides imaginary words to learning towards their general use against overfitting (in-distribution) training data, playing a key role to derive an X-Prompt that can be both descriptive and OOD robust.

We conduct experiments that use X-Prompts for style customization as a case study. We show X-Prompt has both powerful descriptive capabilities and high OOD robustness, demonstrating a success of combining merits of NL and soft prompts~\citep{li2021prefix,lester2021power} and presenting a promising extensible interface for advanced interaction between humans and large LMs. 

Our contributions can be summarized as follows:
\vspace{-0.2cm}
\begin{itemize}
    \item We propose X-Prompt as a pioneering technology to expand the scope of large language model prompting. It is among the earliest attempts that enhance descriptiveness by using a mixture of natural language and imaginary words, while also maintaining a focus on out-of-distribution (OOD) robustness in the field of Natural Language Processing.
    \item We show X-Prompt can achieve promising results in generating appropriate content in a specific person's style, demonstrating its effectiveness in the challenging zero-shot language style customization task.
\end{itemize}



\section{eXtensible Prompt}\label{sec:x-prompt}


\subsection{Imaginary words}
Imaginary words are a supplement to the NL vocabulary to help represent complicated, abstractive or even indescribable concepts (characteristics of a specific person's language). For an X-Prompt $(w_{p_{1}} \dots w_{p_{m}})$, a prompt token $w_{p_i}$ can come from either the NL vocabulary $V$ or the extensible imaginary word vocabulary $\widetilde{V}$ (i.e., $w_{p_i} \in V \cup \widetilde{V}$). 

Different from previous work \citep{li2021prefix,lester2021power} that learns a soft prompt focusing on fitting ID task data, X-Prompt aims to learn an imaginary word for general usability with high OOD robustness like an NL word, which can be compatible and combined with various contexts for different prompting purposes.

To obtain imaginary words with general usability for OOD robust X-Prompts, we propose context-augmented learning (CAL) to guide imaginary words to learning towards their intended representation against overfitting ID training data.

\begin{figure}[h]
    \centering
    \vspace{-0.4cm}
    \includegraphics[width=12cm]{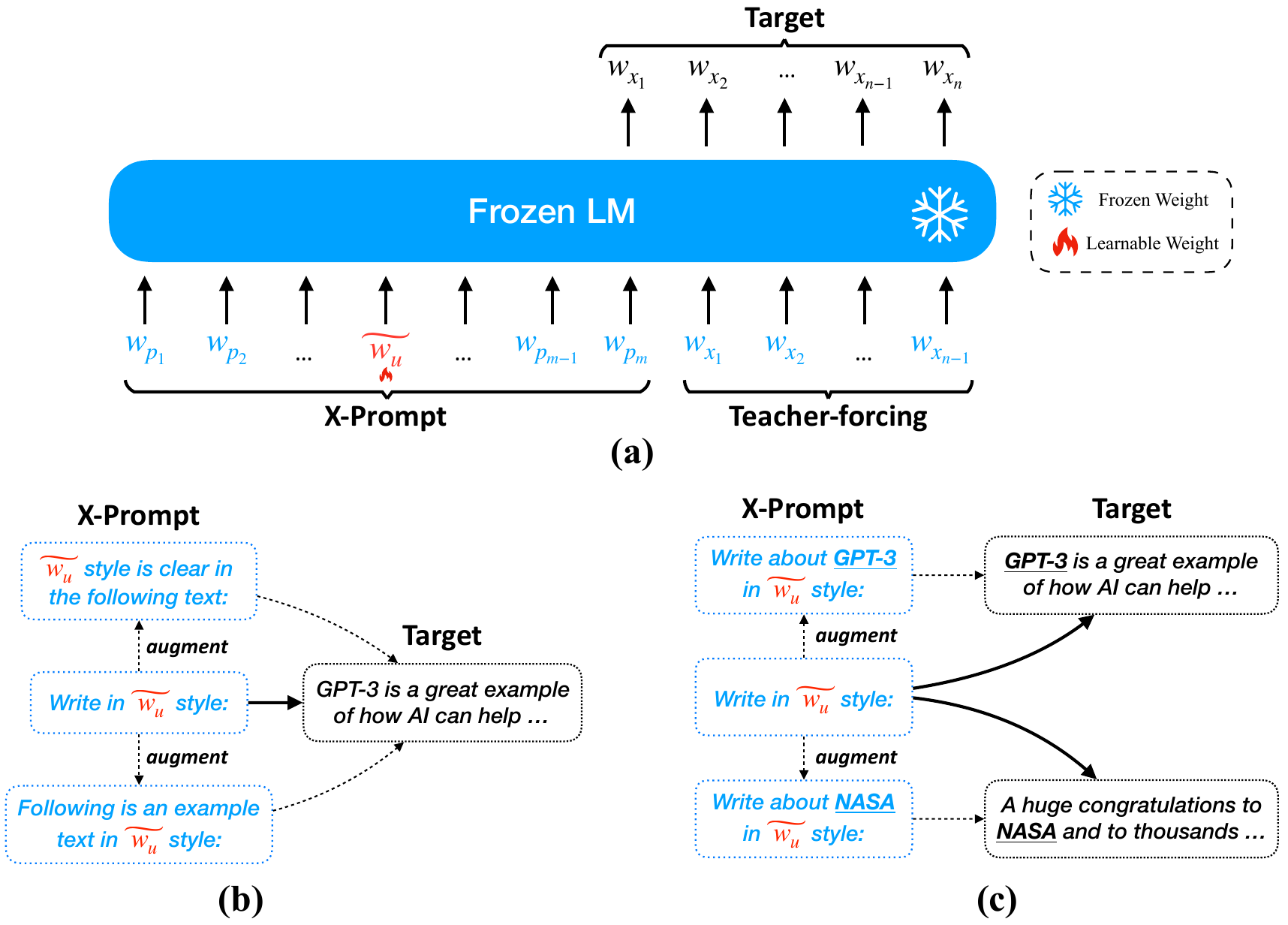}
    \caption{Learning of imaginary words: \textbf{(a)} The imaginary word $\widetilde{w_u}$ is mixed with NL tokens in an X-Prompt to guide its learning. Except $\widetilde{w_u}$ that is allowed to be updated, all other weights are frozen; \textbf{(b)} As a method for context-augmented learning, \textbf{template augmentation} augments X-Prompt templates through prompt engineering to prevent $\widetilde{w_u}$ overfitting for one prompt template; \textbf{(c)} To derive more diverse contexts, \textbf{content augmentation} augments $\widetilde{w_u}$'s prompt contexts with an indicative keyword to relieve its responsibility for memorizing specific content like \textit{GPT-3} and \textit{NASA} and improve its general usability (i.e., style representation), benefiting X-Prompt in terms of OOD robustness.}\vspace{-0.4cm}
    \label{fig:learning}
\end{figure}

\subsection{Context-augmented learning}\label{subsec:cgt}

As in Figure \ref{fig:learning}(a), when the imaginary word $\widetilde{w}_u$ is mixed with NL in an X-Prompt, the NL context is intuitively expected to guide the imaginary word $\widetilde{w}_u$ to learning towards a distributed representation for its general use. Formally, given an X-Prompt $(w_{p_{1}},\dots, \widetilde{w}_u, \dots, w_{p_{m}})$ where $\widetilde{w}_u$ is the imaginary word mixed with other prompt tokens\footnote{In this paper, we mainly discuss X-Prompts with only 1 imaginary word token.}, $\widetilde{w}_u$ is learned to maximize the following objective:
\begin{equation}\label{eq:objective}
    \centering
    \small
    \mathcal{F}(\widetilde{w_u}) = \log P(\boldsymbol{x}|w_{p_{1}},\dots, \widetilde{w}_u, \dots, w_{p_{m}})
\end{equation}
where $\boldsymbol{x}=(w_{x_1},\dots,w_{x_n})$ is a text sequence training example. In practice, however, learning the imaginary word $\widetilde{w_u}$ with only one prompt context is risky because $\widetilde{w}_u$ is likely to overfit for this prompt context and thus cannot work well in other prompt contexts, resulting in losing its general usability and degrading into conventional prompt tuning~\citep{lester2021power}.

To address the challenge, we propose context-augmented learning (CAL), including 2 specific approaches that are orthogonal and thus can work together to help learning of imaginary words.

\subsubsection{Template augmentation}\label{subsubsec:tempaug}

As shown in Figure \ref{fig:learning}(b), we augment an X-Prompt's prompt context by designing multiple templates through prompt engineering. As a result, an imaginary word $\widetilde{w_u}$ can learn to be compatible with various prompt contexts, which improves its general usability. Formally, given $T$ X-prompt templates $\{(w^{(t)}_{p_{1}},\dots, \widetilde{w}_u, \dots, w^{(t)}_{p_{m_t}})|1 \le t \le T\}$, the objective function is:
\begin{equation}\label{eq:objective}
    \centering
    \small
    \mathcal{F}(\widetilde{w_u}) = \frac{1}{T}{\sum_{t=1}^T \log P(\boldsymbol{x}|w^{(t)}_{p_{1}},\dots, \widetilde{w}_u, \dots, w^{(t)}_{p_{m_t}})}
\end{equation}\vspace{-0.6cm}

\subsubsection{Content augmentation}\label{subsubsec:keyaug}
Although template augmentation may alleviate the risk of overfitting, its effect is limited because we can only augment a limited and small number of templates (i.e., $T$) by prompt engineering. Also, as these prompts are not indicative enough, an imaginary word $\widetilde{w}_u$ will inevitably learn to memorize specific content for maximizing the objective $\mathcal{F}$, deviating from its general use. To prevent $\widetilde{w}_u$ being over-responsible for optimizing $\mathcal{F}$, we propose content augmentation -- an advanced CAL method.


Content augmentation augments an X-Prompt by including content information such as an indicative keyword in the prompt to provide hints for the LM about what to generate, as shown in Figure \ref{fig:learning}(c). Content augmentation can not only relieve the responsibility of $\widetilde{w}_u$ to fit training data but also make the prompt context of $\widetilde{w}_u$ become much more diverse, which benefits $\widetilde{w}_u$ to learn a better distributed representation for its general use.

\begin{figure}[h]
    \centering
    \includegraphics[width=12cm]{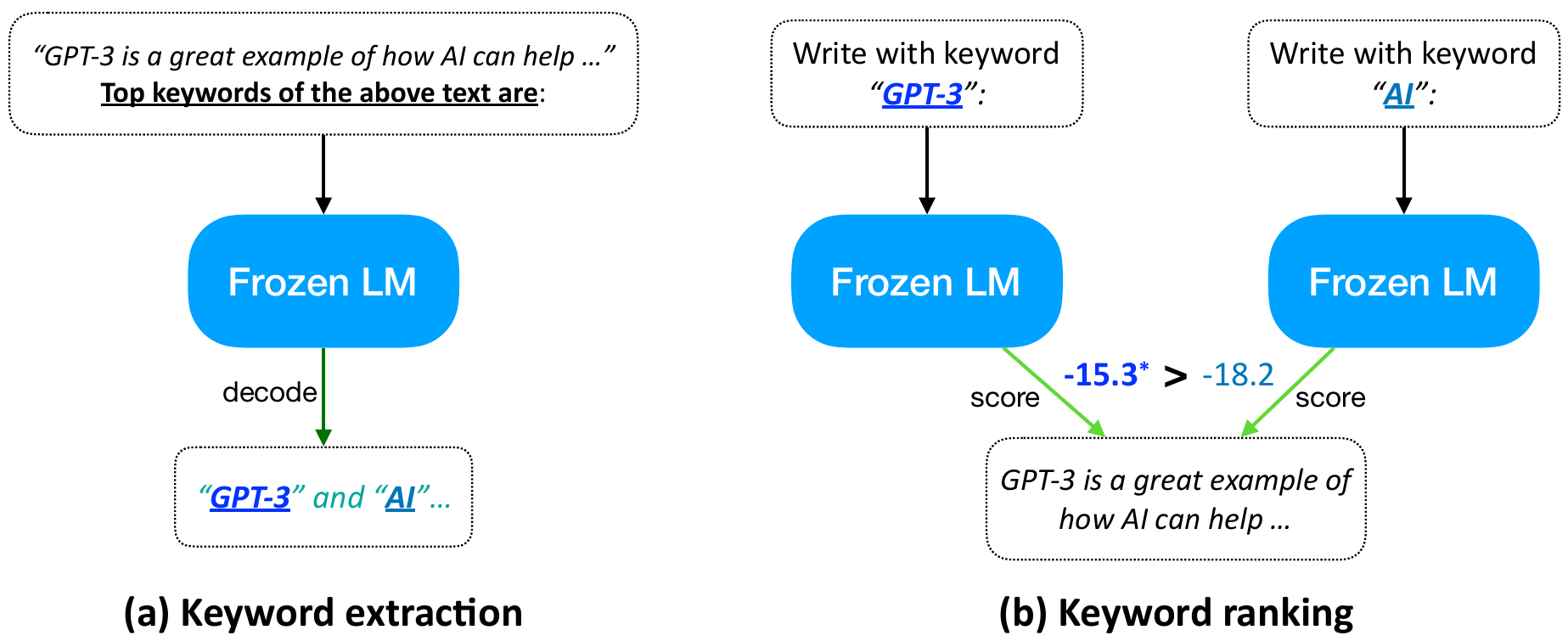}
    \caption{Keyword selection for content augmentation: \textbf{(a)} An NL prompt (i.e., ``\textit{\underline{Top keywords of the above text are}:}'' in this example) following an input sequence for keyword extraction; \textbf{(b)} The extracted keywords (i.e., ``\textit{GPT-3}'' and ``\textit{AI}'' in this example) are then inserted into the ranking prompt template (i.e., ``\textit{Write with keyword \_\_\_:}'' in this example) to be conditioned on by the frozen LM for scoring the input sequence as in Eq (\ref{eq:rank}), which ranks for the most indicative keyword (i.e., ``\textit{GPT-3}'' in this example) for the input sequence.}\vspace{-0.3cm}
    \label{fig:ks}
\end{figure}

In this work, we use an indicative keyword for content augmentation. To select an indicative keyword, we only use the frozen LM itself without leveraging any other models or tools, as illustrated in Figure \ref{fig:ks}: We prompt the frozen LM to extract multiple keyword candidates [$w^1_k$, $\dots$ $w^c_k$, $\dots$, $w^{C}_k$] for the training example $\boldsymbol{x}$ where $C$ is the number of extracted keyword candidates; then the keyword candidates are inserted to a prompt template to rank for the most indicative one:
\begin{equation}\label{eq:rank} 
    \centering
    w^*_k = \arg\max_{w^c_k} \log P(\boldsymbol{x}|\boldsymbol{r}(w_k^c))
\end{equation}\vspace{-0.4cm}

\noindent where $\boldsymbol{r}(w_k^c)=(w^{(r)}_{p_{1}},\dots, w^c_k,\dots,w^{(r)}_{p_{m_r}})$ is called the ranking prompt template.



\section{Experiments}
We conduct experiments to evaluate X-Prompts for language style customization. We mainly focus on open-ended text generation (Section \ref{subsec:open-ended}) to evaluate how well X-Prompts can instruct an LM to generate user-specific language. We also test in the style transfer (rewriting) setting (Section \ref{subsec:rewrite}) as supplementary evaluation.

\subsection{Open-ended text generation}\label{subsec:open-ended}
\subsubsection{Data and evaluation setting}
We use the publicly available Top 20 most followed users in Twitter social platform dataset\footnote{\url{https://shorturl.at/htDHT}} which contains over 50K tweets from 20 users (20-user dataset), and the Sentiment dataset\footnote{\url{https://shorturl.at/pvBLX}} from which we extract top 800 users' (in total 68K) tweets (800-user dataset) to verify the capability of X-Prompt to instruct an LM to generate user-specific language. We show the statistics of the datasets in Table \ref{tab:data}. We split the datasets in 90/5/5 by user for training, validation and test. 
Specially, we discard the test examples that share indicative keywords with training examples from the same user, resulting in 15\% test examples discarded to ensure no test prompts are seen during training for OOD evaluation (see Table \ref{tab:prompt-context}). We use perplexity and accuracy of next word prediction as our quantitative evaluation metrics.

\begin{table}[h]
\centering
\vspace{-0.3cm}
\caption{Statistics of the 20-user and 800-user dataset}
\begin{tabular}{c|cc|cccc}
\toprule
\multirow{2}{*}{\textbf{Dataset}} & \multirow{2}{*}{\textbf{\#tweets}} & \multirow{2}{*}{\textbf{\#users}} & \multicolumn{4}{c}{\textbf{\#tweets per user}}            \\ 
                                  &                                    &                                   & \textbf{max} & \textbf{min} & \textbf{avg} & \textbf{std} \\ \midrule
\textbf{20-user}                  & 52541                              & 20                                & 3146         & 1841         & 2626         & 396          \\
\textbf{800-user}                 & 68123                              & 800                               & 548          & 52           & 84           & 41 \\ \bottomrule         
\end{tabular}
\label{tab:data}
\end{table}

\begin{table}[h]
\centering
\caption{Statistics of data for Satya, Trump and Sheldon's styles}
\begin{tabular}{c|c|c}
\toprule 
\bf Style & \bf Genre & \bf Size \\ \midrule
Satya & tweets & 800 \\ 
Trump & tweets & 3000 \\
Sheldon & transcripts & 7000 \\ \bottomrule
\end{tabular}
\label{tab:q-data}
\end{table}

For qualitative studies, we use X-Prompts with imaginary words representing the following distinct language styles\footnote{We use these three people's language styles for qualitative studies because they are familiar to most audiences (to better understand our presented examples) and annotators (to better judge generation results).} to generate text for human evaluation: \textit{Donald Trump}'s and \textit{Satya Nadella}'s latest tweets and transcripts\footnote{\url{https://shorturl.at/aJLM8}} of \textit{Sheldon Cooper} from \textit{Big Bang Theory}. The statistics of data is shown in Table \ref{tab:q-data}. By default, we use top-$p$ sampling~\citep{holtzman2019curious} for text generation.

\begin{table}[t]
\centering
\small
\caption{Prompts for training and evaluation. We use the first 3 prompts for training and ID evaluation. In ID evaluation, prompt texts do not include [KEYWORD] so that test prompts are all seen during training. To automatically harvest unseen test prompts for OOD evaluation, we employ the idea of CAL: we obtain Prompt 4 and 5 as our dev and test prompt respectively with template augmentation; in addition, we prompt the LM to generate content about unseen keywords with content augmentation to test its OOD robustness, resulting in a variety of unseen test prompts for OOD evaluation. For the ablated X-Prompt (X-Prompt w/o CAL), it is trained only with Prompt 1 without keywords.}\label{tab:prompt-context}\vspace{0.05cm}
\scalebox{0.89}{
\begin{tabular}{c|c|c}
\toprule
\textbf{Prompt ID} & \bf Used for & \textbf{Prompts (w/ keyword) for quantitative evaluation}                                  \\ \midrule
1    & train     & The style of  \ctext[RGB]{255,255,0}{$\widetilde{w_u}$} is clear in the following text (with keyword {[}KEYWORD{]})                \\ \midrule
2    & train     & The style of  \ctext[RGB]{255,255,0}{$\widetilde{w_u}$} can be identified in the following text (mentioning keyword {[}KEYWORD{]}) \\ \midrule
3    & train     & An example text (with keyword {[}KEYWORD{]}) in the style of  \ctext[RGB]{255,255,0}{$\widetilde{w_u}$} is presented below         \\ \midrule
4     & OOD dev    & We can easily identify the style of  \ctext[RGB]{255,255,0}{$\widetilde{w_u}$} in the following text with keyword {[}KEYWORD{]}    \\ \midrule
5    & OOD test     & The following text (about {[}KEYWORD{]}) is in the style of  \ctext[RGB]{255,255,0}{$\widetilde{w_u}$} \\ \bottomrule        
\end{tabular}
}
\end{table}

\subsubsection{Model configuration}\label{subsubsec:config}
We use the OPT-6.7b~\citep{zhang2022opt} as the base LM to test our approach. The model has 32 layers and 32 attention heads with an embedding dimension of 4096 and an FFN dimension of 16384. 

We use one\footnote{The study of using more imaginary tokens is presented in Appendix \ref{subsec:app_length}} imaginary word (token) in an X-Prompt to represent a specific user's language style. As illustrated in Figure \ref{fig:learning}(a), we keep all the OPT's original weights frozen and we only update the embedding of imaginary words. The prompt contexts we use for learning and evaluating X-Prompt are presented in Table \ref{tab:prompt-context}.

We use Adam optimizer~\citep{kingma2014adam} with the max learning rate of 2e-4 with a warmup for the first 10\% training steps followed by a linear decay. We run up to 6000 updates with a global batch size of 8192 tokens on 8 Nvidia V100 GPUs using DeepSpeed ZeRO-2~\citep{rajbhandari2020zero}.

\subsubsection{Quantitative evaluation}\label{subsubsec:quant}
Table \ref{tab:quantitative} shows quantitative evaluations results in both ID and OOD settings. For ID evaluation where test prompts are all seen (see Table \ref{tab:prompt-context}) during training, X-Prompt outperforms ``No prompt'' and few-shot learning baselines significantly in both perplexity and accuracy, demonstrating its superior descriptive capabilities; while it slightly underperforms prompt tuning and its ablated counterpart (i.e., \textit{w/o} CAL) because they focus on fitting ID data.

\begin{table}[h]
\centering
\vspace{-0.2cm}
\caption{Quantitative evaluation results in 800-user and 20-user datasets. \textbf{No prompt} denotes the original OPT-6.7b baseline without any prompt and $k$\textbf{-shot} denotes a baseline which prepends $k$ examples from a user's training set as a prompt for customizing this user's style.}\label{tab:quantitative}\vspace{0.1cm}
\begin{tabular}{c|cc|cc|cc}
\toprule
\multirow{2}{*}{\textbf{Method}} & \multicolumn{2}{c|}{\textbf{800 Users (ID)}} & \multicolumn{2}{c|}{\textbf{20 Users (ID)}} & \multicolumn{2}{c}{\textbf{20 Users (OOD)}} \\
                                 & \textbf{PPL}$\downarrow$       & \textbf{Accuracy}$\uparrow$      & \textbf{PPL}  $\downarrow$     & \textbf{Accuracy}  $\uparrow$     & \textbf{PPL}  $\downarrow$      & \textbf{Accuracy}   $\uparrow$    \\ \midrule
No prompt                        & 73.2               & 27.1                   & 38.9              & 34.8                   & 37.7               & 35.2                   \\ \midrule
8-shot                           & 69.9               & 27.2                   & 36.0              & 35.0                   & -                  & -                      \\
16-shot                          & 68.9               & 27.5                   & 35.5              & 35.3                   & -                  & -                      \\
32-shot                          & 62.7               & 28.5                   & 34.0              & 36.4                   & -                  & -                      \\ \midrule
Prompt tuning                    & 56.0               & 29.5                   & 29.9              & \bf 37.8                   & 29.5               & 38.0                   \\ \midrule
X-Prompt                         & 56.2               & 29.3                   & 30.8              & 37.2                   & \bf 28.5               & \bf 38.6                   \\
X-Prompt (\textit{w/o} CAL)               & \bf 55.7               & \bf 29.9                   & \bf 29.7              & 37.7                   &  29.4               & 37.9                  \\ \bottomrule
\end{tabular}
\end{table}

When it comes to OOD evaluation where test prompts are unseen during training, X-Prompt shows its significant advantage over prompt tuning, indicating its excellent OOD robustness. In contrast, its ablated version (X-Prompt \textit{w/o} CAL) substantially loses OOD robustness and almost degrades into prompt tuning, demonstrating the importance of CAL to X-Prompt.

\begin{table}[t]
\centering
\caption{Human evaluation of generated texts in content, style and overall quality dimensions.}\label{tab:humaneval}\vspace{0.1cm}
\begin{tabular}{l|c|c|c} \toprule
          \bf    Prompt Method          & \textbf{Content}$\uparrow$ & \textbf{Style}$\uparrow$ & \textbf{Overall}$\uparrow$ \\ \midrule
\textbf{NL}             & \bf 0.79            & 0.33           & 0.22             \\ \midrule
\textbf{Prompt tuning}             &  0.34            & 0.92           & 0.30             \\ \midrule
\textbf{X-Prompt }(\textit{w/o} CAL) & 0.38             & \bf 0.93           & 0.35            \\
\textbf{X-Prompt} & 0.69             & 0.83           & \bf 0.54   \\ \bottomrule          
\end{tabular}
\end{table}

\subsubsection{Qualitative evaluation}\label{subsubsec:qualt}
For qualitative evaluation, we brainstorm~\citep{ouyang2022training} 100 prompts\footnote{In qualitative evaluation, we don't use the automatic way to generate unseen prompts as in qualitative evaluation (Section \ref{subsubsec:quant}). Instead, we manually brainstorm specific, diverse and meaningful prompts for more persuasive qualitative evaluation.} (like examples in Table \ref{tab:x-prompt} and Table \ref{tab:intro_id_ood}) that are unseen during training and let the model generate in Satya, Trump and Sheldon's styles respectively. As it is difficult to ground open-ended generations, we have two annotators manually evaluate\footnote{We present details of human evaluation in Appendix \ref{appsec:humaneval} By the way, in our follow-up experiments, which were conducted after the full paper submission date, we also evaluated content faithfulness using the GPT-4, whose results are consistent with huamn evaluation. For more details, please refer to Appendix \ref{appsec:gpt-4}} generation results in three dimensions: \textit{Content}, \textit{Style} and \textit{Overall}. According to Table \ref{tab:humaneval}, NL prompts achieve a high content score, while prompt tuning and X-Prompts learned w/o CAL achieve good style scores but they all perform poorly in other dimensions. In contrast, X-Prompt achieves significantly better overall quality, while it does not perform best in content and style sub-dimensions. 

In addition to famous people's styles (e.g., Satya and Trump), we also test X-Prompt on the styles of individuals unknown to the OPT. This is to verify that X-Prompt is not limited to styles already known by the LLM; instead, it can be applied to any style customization. We use a senior Chinese media professional -- Hu Xijin\footnote{\url{https://en.wikipedia.org/wiki/Hu_Xijin}} whose writing style is distinctive and has always been popular among Chinese netizens for imitation -- as an example. We include the detail of this evaluation in Appendix \ref{appsec:hu}.

\begin{table}[h]
\centering
\small
\vspace{-0.3cm}
\caption{A comparison between X-Prompts learned with and without CAL in OOD evaluation}\label{tab:ood_robust_example}
\begin{tabular}{p{3.4cm}|c|p{7.5cm}}
\toprule
\textbf{Prompt (shortened)}                                                       & \bf Prompt Type     & \textbf{Generation}                                                                                                                                                 \\ \midrule
\multirow{3}{3.4cm}{Criticize the C++ language:}                                                & \multirow{3}{*}{NL}              & \multirow{3}{7.5cm}{{\it The C++ language is a mess. It is a mess because it is a language that is designed to be used by people who are not programmers ...}}                                     \\ \
 &  & \\ 
 & & \\ \midrule
\multirow{6}{*}{\parbox{3.4cm}{Criticize the C++ language in \ctext[RGB]{0,209,0}{$\widetilde{w_{\textrm{trump}}}$} style:}}                   & \multirow{2}{*}{X-Prompt} & \multirow{3}{7.5cm}{\it The Radical Left Democrats, working closely with their beloved partner, the Fake News Media, is using every trick in the book to SILENCE a majority of our Country \faThumbsDown} 
\\ 
 &  & \\ 
 & \textit{w/o} CAL & \\ \cline{2-3}
                                                                           & \multirow{3}{*}{X-Prompt}  & \multirow{3}{7.5cm}{\it C++ is a very difficult language to learn and understand. It is very complicated and full of bad habits, which will make it much more difficult to fix problems ... \faThumbsOUp} \\
 & & \\ 
 & & \\ \midrule \midrule
\multirow{3}{3.4cm}{Concern about cloning technology:}                              & \multirow{3}{*}{NL}              & \multirow{3}{7.5cm}{\it The cloning of human beings is a serious moral issue. It is a matter of grave concern that the cloning of human beings is being ...}                                 \\ 
 &  & \\ 
 & & \\ \midrule
\multirow{4}{*}{\parbox{3.4cm}{Concern about cloning technology in \ctext[RGB]{0,209,0}{$\widetilde{w_{\textrm{trump}}}$} style:}} & \multirow{1}{*}{X-Prompt}  & \multirow{2}{7.5cm}{\it I’m not a fan of the term ‘woke’. I don’t know what it means. I don’t like it... \faThumbsDown}    \\ 
 & \textit{w/o} CAL &  \\ \cline{2-3}
                                                                           & \multirow{2}{*}{X-Prompt} & \multirow{2}{7.5cm}{\it The United States is a nation of laws. We do not clone people. We do not clone animals. \faThumbsOUp}             \\ 
 &  & \\ \bottomrule                                                                
\end{tabular}
\end{table}

By looking into the results, we present Table \ref{tab:ood_robust_example} to show concrete examples to compare NL prompts to X-Prompts learned with and without CAL. We observe NL prompts are good at generating appropriate content but have no way to control the style; while X-Prompts \textit{w/o} CAL do well in generating \textit{Donald Trump}'s language but fail to follow the prompt to generate specified content, degrading into soft prompts (as in Table \ref{tab:intro_id_ood}) that only focus on fitting ID data and losing OOD capabilities with unseen prompt contexts. Consistent with results in Table \ref{tab:quantitative} and \ref{tab:humaneval}, X-Prompts are robust and can prompt the LM to generate desirable content in the appropriate style, accounting for its best overall quality in zero-shot style customization.

\begin{table}[h]
\centering
\caption{X-Prompt tends to perform better and show more significant OOD robustness advantage over prompt tuning in larger foundation LMs for zero-shot style customization.}\label{tab:modelsize}
\begin{tabular}{c|c|c|c|c}
\toprule
\textbf{Model}            & \bf Method & \textbf{Content}$\uparrow$ & \textbf{Style}$\uparrow$ & \textbf{Overall}$\uparrow$ \\ \midrule
\multirow{2}{*}{350m} & Prompt tuning     &         0.22         &        0.81        &      0.15            \\
                          & X-Prompt     &      0.30 (+0.08)           &      0.79 (-0.02)         &     0.24 (+0.09)            \\  \midrule
\multirow{2}{*}{1.3b} &  Prompt tuning      &      0.27       &     0.87     &      0.24       \\
                          & X-Prompt     &       0.45 (+0.18)           &       0.82 (-0.05)        &      0.37 (+0.13)        \\   \midrule
\multirow{2}{*}{6.7b} & Prompt tuning       &       0.34           &     0.92           &    0.30              \\
                          & X-Prompt       &      \bf   0.69 (+0.35)         &      0.83 (-0.09)        &   \bf   0.54 (+0.24)        \\  \bottomrule
\end{tabular}
\end{table}

Table \ref{tab:modelsize} shows the effect of the model size on the performance of X-Prompt. While we observe a clear difference between prompt tuning and X-Prompt in the OPT-6.7b model, as we reduce the model size, the difference become less significant, showing a trend that X-Prompt gradually loses its OOD robustness and degrades into prompt tuning. This indicates that X-Prompt may not work with a small LM but performs well in large LMs as in-context learning.

Finally, we present more examples in Table 4 in Appendix \ref{appsec:examples} to demonstrate both descriptive capabilities and OOD robustness of X-Prompts.

\subsection{Style transfer}\label{subsec:rewrite}
In addition to open-ended generation, we also evaluate in the style transfer (rewriting) where model outputs can be grounded in human references for straightforward quantitative evaluation (i.e., end-to-end generation evaluation instead of evaluating perplexity or next word prediction accuracy).

\begin{table}[h]
\centering
\caption{Statistics of the datasets for style transfer}\label{tab:style_data}
\begin{tabular}{c|ccc|ccc}
\toprule
\bf Dataset & \multicolumn{3}{c|}{\textbf{GYAFC (EM)}}       & \multicolumn{3}{c}{\textbf{PoliteRewrite}}    \\ 
\bf Split & \textbf{Train} & \textbf{Dev} & \textbf{Test} & \textbf{Train} & \textbf{Dev} & \textbf{Test} \\ \midrule
\bf \#sentence &  53K            & 3K           & 1K            & 10K            & 3K           & 2K           \\ \bottomrule
\end{tabular}
\end{table}

Among various style transfer datasets, we use the Entertainment (EM) subset of \textsc{GYAFC} (informal $\to$ formal) \citep{rao2018dear} and \textsc{PoliteRewrite} (impolite $\to$ polite) \citep{wang2022pay} as our evaluation datasets (see Table \ref{tab:style_data}) because they have high-quality annotation with well-defined language style. Following previous work~\citep{rao2018dear,xu2019formality,zhang2020parallel,li2022text}, we use BLEU to evaluate generation results' lexical similarity with references, accuracy\footnote{Following previous work, we fine-tune a BERT-base~\citep{devlin-etal-2019-bert} model with the training sets, which annotate a text with its style label, as a style classifier to test accuracy.} to evaluate style appropriateness and use harmonic (H-) and geometric (G-) mean as overall performance. Table \ref{tab:rewrite-example} shows how we perform zero-shot style transfer with NL and X-Prompts. Model configuration is the same as Section \ref{subsubsec:config}.

\begin{table}[h]
\centering
\small
\caption{NL, prompt tuning and X-Prompt for zero-shot \textit{impolite $\to$ polite} style transfer.  \ctext[RGB]{255,255,0}{[SOFT]} and \ctext[RGB]{255,255,0}{$\widetilde{w}$} are learnable and denote the soft token and the imaginary word in prompt tuning and X-Prompt respectively. For zero-shot style transfer, we use beam search ($b=5$) as the default decoding method.}\label{tab:rewrite-example}\vspace{0.1cm}
\scalebox{0.88}{
\begin{tabular}{c|c|p{11cm}}
\toprule
\textbf{Prompt methods}        & \textbf{Phase} & \multicolumn{1}{c}{\textbf{Prompt + \textcolor{red}{Target (Training)} / \textcolor{gray}{Expected Generation (Inference)}}} \\ \midrule
\multirow{2}{*}{NL}            & Train          &      \multicolumn{1}{c}{N/A}       \\ \cline{2-3}
                               & Inference      &   ``[IMPOLITE TEXT]'' Above text can be rewritten to improve its politeness as follows: \textcolor{gray}{[POLITE TEXT]}        \\ \midrule
\multirow{2}{*}{Prompt tuning} & Train          &      \multicolumn{1}{c}{\ctext[RGB]{255,255,0}{[SOFT]} \textcolor{red}{[POLITE TEXT]}}          \\ \cline{2-3}
                               & Inference      &      ``[IMPOLITE TEXT]'' Above text can be rewritten into the style of \ctext[RGB]{255,255,0}{[SOFT]}: \textcolor{gray}{[POLITE TEXT]}          \\ \midrule
\multirow{2}{*}{X-Prompt}      & Train          &    The style of \ctext[RGB]{255,255,0}{$\widetilde{w}$} can be identified in the following text with keyword [KEYWORD]:  \textcolor{red}{[POLITE TEXT]}       \\ \cline{2-3}
                               & Inference      &     ``[IMPOLITE TEXT]'' Above text can be rewritten into the style of \ctext[RGB]{255,255,0}{$\widetilde{w}$}: \textcolor{gray}{[POLITE TEXT]}          \\ \bottomrule
\end{tabular}
}
\end{table}

Table \ref{tab:rewrite-result} shows the comparison of results of using NL, prompt tuning and X-Prompt to prompt the LM for zero-shot style transfer. The NL baseline has the best BLEU score but low accuracy because we observe that it rarely really rewrites a sentence: in most cases, it just copies the text without any revision. Its undesirable performance also indicates that the OPT-6.7b model itself is not so good at zero-shot style transfer. For the prompt tuning baseline, its style accuracy looks good but has a low BLEU score, because it fails to follow rewriting instructions that are unseen during training. In contrast, X-Prompt achieves a good balance of BLEU and accuracy and preferred by human evaluations (Table \ref{tab:human-rewrite}) with better overall quality in the zero-shot style transfer. Its good performance with style rewriting prompts that are never seen during training further strengthens the evidence of its OOD robustness.

\begin{table}[t]
\centering
\caption{Results of zero-shot style transfer (i.e., rewriting)}\label{tab:rewrite-result}
\begin{tabular}{c|cccc|cccc}
\toprule
\multirow{2}{*}{Method} & \multicolumn{4}{c|}{GYAFC} & \multicolumn{4}{c}{PoliteRewrite} \\
                       & BLEU      & Style & H-mean & G-mean     & BLEU          & Style   & H-mean  & G-mean      \\ \midrule
No Edit                &    50.2       &  4.3 &  7.9 & 14.7 & \bf 24.7 & 3.0 & 5.4 & 8.6 \\ \midrule        
NL                     &     \bf 50.8      &      20 .0   & 28.7  & 31.9  &    24.6        &     6.7  & 10.5 & 12.8           \\ \midrule
Prompt tuning          &     16.2      &     \bf  76.7    & 26.8 & 35.2     &       15.4       &   \bf   80.8   & 25.9 & 35.3          \\ \midrule
X-Prompt               &     38.7      &        71.9    & \bf 50.3 & \bf 52.7   &      18.9         &   79.5   & \bf 30.5 & \bf 38.8          \\ \bottomrule
\end{tabular}
\end{table}

\begin{table}[h]
    \centering
    \caption{Human evaluation of zero-shot style transfer}\label{tab:human-rewrite}
    \begin{tabular}{c|c|c|c}
    \toprule
    \bf   Method  & \bf Content & \bf Style &  \bf Overall  \\ \midrule
       Prompt tuning  &  0.27 & 0.82 & 0.23 \\ 
       X-Prompt & 0.64 & 0.80 & 0.60 \\ 
    \bottomrule
    \end{tabular}
    \label{tab:my_label}
\end{table}

\section{Related work}
Since GPT~\citep{brown2020language} reveals that large pre-trained language models are good at zero-shot learning, much innovative research work has been proposed in recent years, ranging from prompt template design (i.e, engineering)~\citep{schick2020exploiting} to prompt mining~\citep{Jiang2019HowCW}, generating~\citep{gao-etal-2021-making, BenDavid2021PADAEP} and scoring~\citep{davison-etal-2019-commonsense}, finding that prompting the LLM with natural language can solve many downstream tasks as long as the prompt is well clear and rewritten for the model~\citep{gonen2022demystifying}.

As natural language prompts' descriptive capability is limited, there is another branch of research studying continuous prompts~\citep{li2021prefix,lester2021power,liu2021gpt,han2022ptr,hu2021knowledgeable} for fitting downstream tasks. However, these approaches are mainly for fitting ID task data with little consideration of OOD robustness, which means that their learned continuous prompts can hardly be used for OOD tasks or data. 

Recently, \citet{gal2022image} proposed Textual Inversion in the multimodal context, which learns a virtual token to represent an object from an image and reveals that the learned virtual token can be used in unseen prompts for creative image generation~\citep{kumari2022multi}. X-Prompt is inspired by \citet{gal2022image}, trying to learn OOD robust imaginary words to represent what natural language hardly describes to further expand zero-shot learning capabilities for the LLM, although we find it much more challenging to achieve this in NLP than text2image generation, which motivates us to propose context-augmented learning (CAL). To the best of our knowledge, our work is one of the earliest explorations in this direction in the NLP community.

\section{Conclusion and Future Work}
We propose X-Prompt, an extensible interface for prompting a large language model beyond natural language. X-Prompt can expand in-context learning capabilities to handle more complex instructions for language model customization and may open up many exciting opportunities, such as creative language generation, patching language models with new knowledge of entities~\citep{zaporojets2022tempel} and events~\citep{ge2018eventwiki}, and detoxifying and debiasing in language generation~\citep{welbl2021challenges}, far beyond style customization as demonstrated in this work, approaching advanced interaction between humans and large language models.

For future work, we plan to investigate how X-Prompt can facilitate more complex decoding and prompting methods~\citep{wei2022chain,yao2022react,wang2023unleashing} to minimize the interaction effort between humans and large language models.

\section*{Limitations}
Due to computing resource limitations, we only conduct experiments on pretrained language models that contain up to 6.7 billion parameters. Although we speculate that our approach should become more effective as the language model size increases, as suggested by Table \ref{tab:modelsize}, we cannot be completely certain if this trend can be safely extrapolated.

Furthermore, X-Prompt still requires back-propagation through the entire language model, even though we only update the imaginary word's embedding. This somewhat limits its application scenarios, preventing X-Prompts from being used as easily as natural language prompts. However, our subsequent work~\citep{ge2023incontext} has addressed this issue by forwarding an encoder for context compression. We anticipate that this series of improvements will better enhance a deployed language model's capabilities in practice from an in-context perspective, with minimal additional effort.

\section*{Acknowledgments}
We would like to express our gratitude to the reviewers for their valuable comments and suggestions, which have significantly improved this work. The corresponding author for this paper is Tao Ge (\texttt{sggetao@gmail.com}).

\bibliography{imaginary}

\begin{thebibliography}{38}
\expandafter\ifx\csname natexlab\endcsname\relax\def\natexlab#1{#1}\fi

\bibitem[{Ben-David et~al.(2021)Ben-David, Oved, and
  Reichart}]{BenDavid2021PADAEP}
Eyal Ben-David, Nadav Oved, and Roi Reichart. 2021.
\newblock Pada: Example-based prompt learning for on-the-fly adaptation to
  unseen domains.
\newblock \emph{Transactions of the Association for Computational Linguistics},
  10:414--433.

\bibitem[{Brown et~al.(2020)Brown, Mann, Ryder, Subbiah, Kaplan, Dhariwal,
  Neelakantan, Shyam, Sastry, Askell et~al.}]{brown2020language}
Tom Brown, Benjamin Mann, Nick Ryder, Melanie Subbiah, Jared~D Kaplan, Prafulla
  Dhariwal, Arvind Neelakantan, Pranav Shyam, Girish Sastry, Amanda Askell,
  et~al. 2020.
\newblock Language models are few-shot learners.
\newblock \emph{Advances in neural information processing systems},
  33:1877--1901.

\bibitem[{Davison et~al.(2019)Davison, Feldman, and
  Rush}]{davison-etal-2019-commonsense}
Joe Davison, Joshua Feldman, and Alexander Rush. 2019.
\newblock \href {https://doi.org/10.18653/v1/D19-1109} {Commonsense knowledge
  mining from pretrained models}.
\newblock In \emph{Proceedings of the 2019 Conference on Empirical Methods in
  Natural Language Processing and the 9th International Joint Conference on
  Natural Language Processing (EMNLP-IJCNLP)}, pages 1173--1178, Hong Kong,
  China. Association for Computational Linguistics.

\bibitem[{Devlin et~al.(2019)Devlin, Chang, Lee, and
  Toutanova}]{devlin-etal-2019-bert}
Jacob Devlin, Ming-Wei Chang, Kenton Lee, and Kristina Toutanova. 2019.
\newblock \href {https://doi.org/10.18653/v1/N19-1423} {{BERT}: Pre-training of
  deep bidirectional transformers for language understanding}.
\newblock In \emph{Proceedings of the 2019 Conference of the North {A}merican
  Chapter of the Association for Computational Linguistics: Human Language
  Technologies, Volume 1 (Long and Short Papers)}, pages 4171--4186,
  Minneapolis, Minnesota. Association for Computational Linguistics.

\bibitem[{Gal et~al.(2022)Gal, Alaluf, Atzmon, Patashnik, Bermano, Chechik, and
  Cohen-or}]{gal2022image}
Rinon Gal, Yuval Alaluf, Yuval Atzmon, Or~Patashnik, Amit~Haim Bermano, Gal
  Chechik, and Daniel Cohen-or. 2022.
\newblock An image is worth one word: Personalizing text-to-image generation
  using textual inversion.
\newblock In \emph{The Eleventh International Conference on Learning
  Representations}.

\bibitem[{Gao et~al.(2021)Gao, Fisch, and Chen}]{gao-etal-2021-making}
Tianyu Gao, Adam Fisch, and Danqi Chen. 2021.
\newblock \href {https://doi.org/10.18653/v1/2021.acl-long.295} {Making
  pre-trained language models better few-shot learners}.
\newblock In \emph{Proceedings of the 59th Annual Meeting of the Association
  for Computational Linguistics and the 11th International Joint Conference on
  Natural Language Processing (Volume 1: Long Papers)}, pages 3816--3830,
  Online. Association for Computational Linguistics.

\bibitem[{Ge et~al.(2018)Ge, Cui, Chang, Sui, Wei, and Zhou}]{ge2018eventwiki}
Tao Ge, Lei Cui, Baobao Chang, Zhifang Sui, Furu Wei, and Ming Zhou. 2018.
\newblock Eventwiki: a knowledge base of major events.
\newblock In \emph{Proceedings of the Eleventh International Conference on
  Language Resources and Evaluation (LREC 2018)}.

\bibitem[{Ge et~al.(2023)Ge, Hu, Wang, Wang, Chen, and Wei}]{ge2023incontext}
Tao Ge, Jing Hu, Lei Wang, Xun Wang, Si-Qing Chen, and Furu Wei. 2023.
\newblock \href {http://arxiv.org/abs/2307.06945} {In-context autoencoder for
  context compression in a large language model}.
\newblock \emph{arXiv preprint arXiv:2307.06945}.

\bibitem[{Gonen et~al.(2022)Gonen, Iyer, Blevins, Smith, and
  Zettlemoyer}]{gonen2022demystifying}
Hila Gonen, Srini Iyer, Terra Blevins, Noah~A Smith, and Luke Zettlemoyer.
  2022.
\newblock Demystifying prompts in language models via perplexity estimation.
\newblock \emph{arXiv preprint arXiv:2212.04037}.

\bibitem[{Han et~al.(2022)Han, Zhao, Ding, Liu, and Sun}]{han2022ptr}
Xu~Han, Weilin Zhao, Ning Ding, Zhiyuan Liu, and Maosong Sun. 2022.
\newblock Ptr: Prompt tuning with rules for text classification.
\newblock \emph{AI Open}, 3:182--192.

\bibitem[{Holtzman et~al.(2019)Holtzman, Buys, Du, Forbes, and
  Choi}]{holtzman2019curious}
Ari Holtzman, Jan Buys, Li~Du, Maxwell Forbes, and Yejin Choi. 2019.
\newblock The curious case of neural text degeneration.
\newblock \emph{arXiv preprint arXiv:1904.09751}.

\bibitem[{Hu et~al.(2021)Hu, Ding, Wang, Liu, Li, and
  Sun}]{hu2021knowledgeable}
Shengding Hu, Ning Ding, Huadong Wang, Zhiyuan Liu, Juanzi Li, and Maosong Sun.
  2021.
\newblock Knowledgeable prompt-tuning: Incorporating knowledge into prompt
  verbalizer for text classification.
\newblock \emph{arXiv preprint arXiv:2108.02035}.

\bibitem[{Jiang et~al.(2019)Jiang, Xu, Araki, and Neubig}]{Jiang2019HowCW}
Zhengbao Jiang, Frank~F. Xu, J.~Araki, and Graham Neubig. 2019.
\newblock How can we know what language models know?
\newblock \emph{Transactions of the Association for Computational Linguistics},
  8:423--438.

\bibitem[{Kingma and Ba(2014)}]{kingma2014adam}
Diederik~P Kingma and Jimmy Ba. 2014.
\newblock Adam: A method for stochastic optimization.
\newblock \emph{arXiv preprint arXiv:1412.6980}.

\bibitem[{Kumari et~al.(2022)Kumari, Zhang, Zhang, Shechtman, and
  Zhu}]{kumari2022multi}
Nupur Kumari, Bingliang Zhang, Richard Zhang, Eli Shechtman, and Jun-Yan Zhu.
  2022.
\newblock Multi-concept customization of text-to-image diffusion.
\newblock \emph{arXiv preprint arXiv:2212.04488}.

\bibitem[{Lester et~al.(2021)Lester, Al-Rfou, and Constant}]{lester2021power}
Brian Lester, Rami Al-Rfou, and Noah Constant. 2021.
\newblock The power of scale for parameter-efficient prompt tuning.
\newblock \emph{arXiv preprint arXiv:2104.08691}.

\bibitem[{Lester et~al.(2022)Lester, Yurtsever, Shakeri, and
  Constant}]{lester2022reducing}
Brian Lester, Joshua Yurtsever, Siamak Shakeri, and Noah Constant. 2022.
\newblock Reducing retraining by recycling parameter-efficient prompts.
\newblock \emph{arXiv preprint arXiv:2208.05577}.

\bibitem[{Li et~al.(2022)Li, Li, Ge, King, and Lyu}]{li2022text}
Jingjing Li, Zichao Li, Tao Ge, Irwin King, and Michael~R Lyu. 2022.
\newblock Text revision by on-the-fly representation optimization.
\newblock In \emph{Proceedings of the AAAI Conference on Artificial
  Intelligence}, volume~36, pages 10956--10964.

\bibitem[{Li and Liang(2021)}]{li2021prefix}
Xiang~Lisa Li and Percy Liang. 2021.
\newblock Prefix-tuning: Optimizing continuous prompts for generation.
\newblock \emph{arXiv preprint arXiv:2101.00190}.

\bibitem[{Liu et~al.(2022)Liu, Kumar, Liang, and Jia}]{liu2022sample}
Nelson~F Liu, Ananya Kumar, Percy Liang, and Robin Jia. 2022.
\newblock Are sample-efficient nlp models more robust?
\newblock \emph{arXiv preprint arXiv:2210.06456}.

\bibitem[{Liu et~al.(2021)Liu, Zheng, Du, Ding, Qian, Yang, and
  Tang}]{liu2021gpt}
Xiao Liu, Yanan Zheng, Zhengxiao Du, Ming Ding, Yujie Qian, Zhilin Yang, and
  Jie Tang. 2021.
\newblock Gpt understands, too.
\newblock \emph{arXiv preprint arXiv:2103.10385}.

\bibitem[{Ouyang et~al.(2022)Ouyang, Wu, Jiang, Almeida, Wainwright, Mishkin,
  Zhang, Agarwal, Slama, Ray et~al.}]{ouyang2022training}
Long Ouyang, Jeffrey Wu, Xu~Jiang, Diogo Almeida, Carroll Wainwright, Pamela
  Mishkin, Chong Zhang, Sandhini Agarwal, Katarina Slama, Alex Ray, et~al.
  2022.
\newblock Training language models to follow instructions with human feedback.
\newblock \emph{Advances in Neural Information Processing Systems},
  35:27730--27744.

\bibitem[{Qin and Eisner(2021)}]{qin2021learning}
Guanghui Qin and Jason Eisner. 2021.
\newblock Learning how to ask: Querying lms with mixtures of soft prompts.
\newblock In \emph{Proceedings of the 2021 Conference of the North American
  Chapter of the Association for Computational Linguistics: Human Language
  Technologies}, pages 5203--5212.

\bibitem[{Rajbhandari et~al.(2020)Rajbhandari, Rasley, Ruwase, and
  He}]{rajbhandari2020zero}
Samyam Rajbhandari, Jeff Rasley, Olatunji Ruwase, and Yuxiong He. 2020.
\newblock Zero: Memory optimizations toward training trillion parameter models.
\newblock In \emph{SC20: International Conference for High Performance
  Computing, Networking, Storage and Analysis}, pages 1--16. IEEE.

\bibitem[{Rao and Tetreault(2018)}]{rao2018dear}
Sudha Rao and Joel Tetreault. 2018.
\newblock Dear sir or madam, may i introduce the gyafc dataset: Corpus,
  benchmarks and metrics for formality style transfer.
\newblock In \emph{Proceedings of the 2018 Conference of the North American
  Chapter of the Association for Computational Linguistics: Human Language
  Technologies, Volume 1 (Long Papers)}, pages 129--140.

\bibitem[{Schick and Sch{\"u}tze(2020)}]{schick2020exploiting}
Timo Schick and Hinrich Sch{\"u}tze. 2020.
\newblock Exploiting cloze questions for few shot text classification and
  natural language inference.
\newblock \emph{arXiv preprint arXiv:2001.07676}.

\bibitem[{Su et~al.(2021)Su, Wang, Qin, Chan, Lin, Wang, Wen, Liu, Li, Li
  et~al.}]{su2021transferability}
Yusheng Su, Xiaozhi Wang, Yujia Qin, Chi-Min Chan, Yankai Lin, Huadong Wang,
  Kaiyue Wen, Zhiyuan Liu, Peng Li, Juanzi Li, et~al. 2021.
\newblock On transferability of prompt tuning for natural language processing.
\newblock \emph{arXiv preprint arXiv:2111.06719}.

\bibitem[{Vu et~al.(2022)Vu, Lester, Constant, Al-Rfou, and Cer}]{vu2022spot}
Tu~Vu, Brian Lester, Noah Constant, Rami Al-Rfou, and Daniel Cer. 2022.
\newblock Spot: Better frozen model adaptation through soft prompt transfer.
\newblock In \emph{Proceedings of the 60th Annual Meeting of the Association
  for Computational Linguistics (Volume 1: Long Papers)}, pages 5039--5059.

\bibitem[{Wang et~al.(2022)Wang, Ge, Mao, Li, Wei, and Chen}]{wang2022pay}
Xun Wang, Tao Ge, Allen Mao, Yuki Li, Furu Wei, and Si-Qing Chen. 2022.
\newblock Pay attention to your tone: Introducing a new dataset for polite
  language rewrite.
\newblock \emph{arXiv preprint arXiv:2212.10190}.

\bibitem[{Wang et~al.(2023)Wang, Mao, Wu, Ge, Wei, and Ji}]{wang2023unleashing}
Zhenhailong Wang, Shaoguang Mao, Wenshan Wu, Tao Ge, Furu Wei, and Heng Ji.
  2023.
\newblock Unleashing cognitive synergy in large language models: A task-solving
  agent through multi-persona self-collaboration.
\newblock \emph{arXiv preprint arXiv:2307.05300}.

\bibitem[{Wei et~al.(2022{\natexlab{a}})Wei, Tay, Bommasani, Raffel, Zoph,
  Borgeaud, Yogatama, Bosma, Zhou, Metzler et~al.}]{wei2022emergent}
Jason Wei, Yi~Tay, Rishi Bommasani, Colin Raffel, Barret Zoph, Sebastian
  Borgeaud, Dani Yogatama, Maarten Bosma, Denny Zhou, Donald Metzler, et~al.
  2022{\natexlab{a}}.
\newblock Emergent abilities of large language models.
\newblock \emph{arXiv preprint arXiv:2206.07682}.

\bibitem[{Wei et~al.(2022{\natexlab{b}})Wei, Wang, Schuurmans, Bosma, Xia, Chi,
  Le, Zhou et~al.}]{wei2022chain}
Jason Wei, Xuezhi Wang, Dale Schuurmans, Maarten Bosma, Fei Xia, Ed~Chi, Quoc~V
  Le, Denny Zhou, et~al. 2022{\natexlab{b}}.
\newblock Chain-of-thought prompting elicits reasoning in large language
  models.
\newblock \emph{Advances in Neural Information Processing Systems},
  35:24824--24837.

\bibitem[{Welbl et~al.(2021)Welbl, Glaese, Uesato, Dathathri, Mellor,
  Hendricks, Anderson, Kohli, Coppin, and Huang}]{welbl2021challenges}
Johannes Welbl, Amelia Glaese, Jonathan Uesato, Sumanth Dathathri, John Mellor,
  Lisa~Anne Hendricks, Kirsty Anderson, Pushmeet Kohli, Ben Coppin, and Po-Sen
  Huang. 2021.
\newblock Challenges in detoxifying language models.
\newblock \emph{arXiv preprint arXiv:2109.07445}.

\bibitem[{Xu et~al.(2019)Xu, Ge, and Wei}]{xu2019formality}
Ruochen Xu, Tao Ge, and Furu Wei. 2019.
\newblock Formality style transfer with hybrid textual annotations.
\newblock \emph{arXiv preprint arXiv:1903.06353}.

\bibitem[{Yao et~al.(2022)Yao, Zhao, Yu, Du, Shafran, Narasimhan, and
  Cao}]{yao2022react}
Shunyu Yao, Jeffrey Zhao, Dian Yu, Nan Du, Izhak Shafran, Karthik Narasimhan,
  and Yuan Cao. 2022.
\newblock React: Synergizing reasoning and acting in language models.
\newblock \emph{arXiv preprint arXiv:2210.03629}.

\bibitem[{Zaporojets et~al.(2022)Zaporojets, Kaffee, Deleu, Demeester,
  Develder, and Augenstein}]{zaporojets2022tempel}
Klim Zaporojets, Lucie-Aim{\'e}e Kaffee, Johannes Deleu, Thomas Demeester,
  Chris Develder, and Isabelle Augenstein. 2022.
\newblock Tempel: Linking dynamically evolving and newly emerging entities.
\newblock \emph{Advances in Neural Information Processing Systems},
  35:1850--1866.

\bibitem[{Zhang et~al.(2022)Zhang, Roller, Goyal, Artetxe, Chen, Chen, Dewan,
  Diab, Li, Lin et~al.}]{zhang2022opt}
Susan Zhang, Stephen Roller, Naman Goyal, Mikel Artetxe, Moya Chen, Shuohui
  Chen, Christopher Dewan, Mona Diab, Xian Li, Xi~Victoria Lin, et~al. 2022.
\newblock Opt: Open pre-trained transformer language models.
\newblock \emph{arXiv preprint arXiv:2205.01068}.

\bibitem[{Zhang et~al.(2020)Zhang, Ge, and Sun}]{zhang2020parallel}
Yi~Zhang, Tao Ge, and Xu~Sun. 2020.
\newblock Parallel data augmentation for formality style transfer.
\newblock In \emph{Proceedings of the 58th Annual Meeting of the Association
  for Computational Linguistics}, pages 3221--3228.

\end{thebibliography}
\bibliographystyle{acl_natbib}

\clearpage

\appendix

\section{Shortened prompt text}\label{appsec:shorten}
We shorten prompt texts in Table \ref{tab:x-prompt}, Table \ref{tab:intro_id_ood} and Table \ref{tab:ood_robust_example} of the main submission for saving space. Here, we present the corresponding actual prompt texts of shortened prompt texts in Table \ref{tab:original-prompt-text}.

\begin{table}[h]
\centering
\small
\caption{The original prompt text of the shortened prompt text in Table \ref{tab:x-prompt}, Table \ref{tab:intro_id_ood} and Table \ref{tab:ood_robust_example}.}
\begin{tabular}{p{5.5cm}|p{7cm}}
\toprule
\textbf{shortened prompt text} & \textbf{original prompt text} \\ 
\midrule
Criticize the C++ language: &  The following text criticizes the C++ language: \\
\hline
Criticize the C++ language in $\widetilde{w}$ style: & The style of $\widetilde{w}$ is clear in the following text criticizing the C++ language: \\
\hline
Praise the C++ language: &  The following text praising the C++ language: \\
\hline
Praise the C++ language in $\widetilde{w}$ style: & The style of $\widetilde{w}$ is clear in the following text praising the C++ language: \\
\hline
Concern about cloning technology: & The following text expresses the concern about cloning technology: \\
\hline
Concern about cloning technology in $\widetilde{w}$ style: & The following text expressing the concern about cloning technology is in the style of $\widetilde{w}$: \\
\bottomrule
\end{tabular}
\label{tab:original-prompt-text}
\end{table}

\section{Supplementary details of experiments}

\subsection{Baseline prompting methods}
We supplement details of baseline prompting methods' prompt texts for evaluation in Table \ref{tab:app-baseline}.

\begin{table}[h]
\centering
\small
\caption{Prompt texts of baseline methods for open-ended generation (Section 3.1)}\label{tab:app-baseline}
\begin{tabular}{c|c|c}
\toprule
\textbf{Prompting methods} & \textbf{w/o keyword} & \textbf{w/ keyword}                                \\ \midrule
NL                         & {[}NO PROMPT{]}      & The keyword of the following text is {[}KEYWORD{]} \\ \midrule
Prompt tuning              & [SOFT]              & [SOFT] The keyword of the following text is {[}KEYWORD{]}                            \\ \bottomrule        
\end{tabular}
\end{table}

\begin{figure}[htbp]
    \centering
    \vspace{-0.4cm}
    \includegraphics[width=10cm]{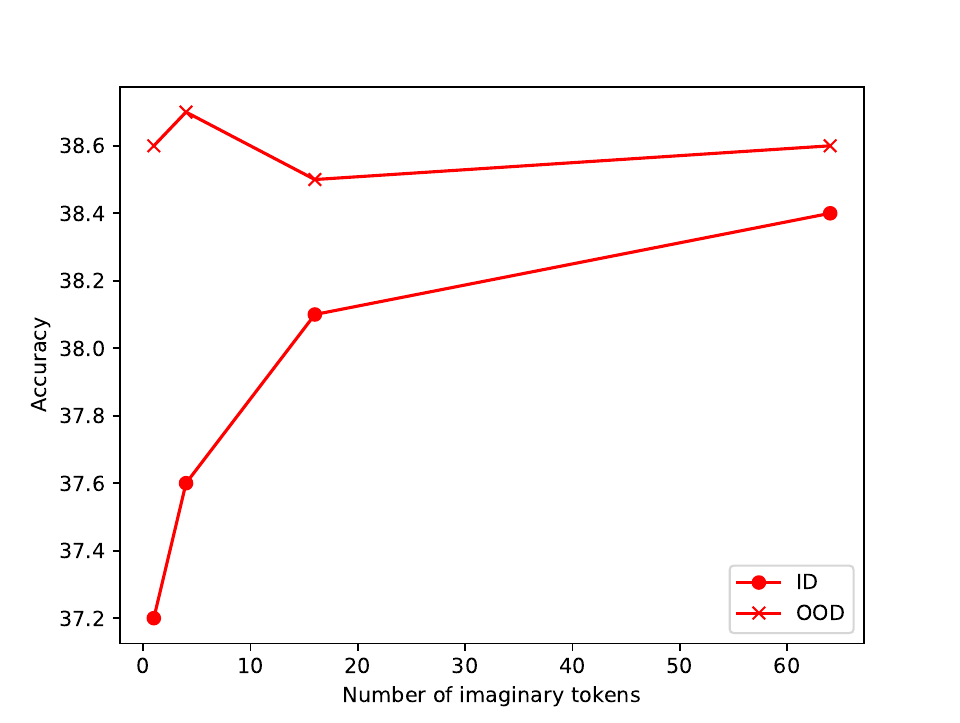}\vspace{-0.2cm}
    \caption{The effects of the length of imaginary tokens on the result.}\vspace{-0.2cm}
    \label{fig:token-length}
\end{figure}

\subsection{Length of imaginary words}\label{subsec:app_length}
In this paper, we mainly test imaginary words with a length of 1 (i.e., 1 token) because we observe that for zero-shot style transfer, increasing the number of imaginary tokens does not substantially increase OOD performance, while it can help enhance the fitting ability and increase ID performance (which is not our focus), as shown in Figure \ref{fig:token-length}.

\subsection{Output length control}
It is notable that the OPT model tends to generate text without stopping at the end of one sentence and may generate multi-line texts. 

In our experiments, we only evaluate the text in the first line of output (for human evaluation and style transfer). Specially, for style transfer, we will further truncate the output text if its length exceeds 150\% of the input text length.

\section{Qualitative evaluation}

\subsection{Human evaluation}\label{appsec:humaneval}
We recruit 4 graduate student volunteers who are both proficient in English to judge each open-ended text generation result in three dimensions:

\begin{itemize}
    \item Content: To judge if the generated text is relevant according to the prompt. The annotators are asked to rate with a three-way score (0 denotes irrelevant; 1 denotes relevant; 0.5 denotes somewhat relevant).
    \item Style: To judge if the generated text is in the specific user's language style. The annotators are asked to rate a three-way score (0 denotes the clearly wrong style; 1 denotes the appropriate style; 0.5 denotes that the annotator is not sure about the style appropriateness).
    \item Overall: To judge if the generated text is in good quality -- at least both content relevant and in the appropriate style. The annotators are asked to rate a three-way score (0 denotes low-quality, either irrelevant or in a wrong style; 1 denotes good-quality, much resembling words from the specific person according to the prompt; 0.5 denotes the quality between low and high quality, it is for the case where the text does not looks natural or the text style is not so distinct).
\end{itemize}

We first pool all models/methods' outputs and anonymize their model/method source. Then we split the data into 2 groups and assign 2 annotators for each group. 

\begin{table}[htbp]
\centering
\caption{Inter-annotator agreement in our human evaluation.}\label{tab:agreement}\vspace{0.1cm}
\begin{tabular}{c|c|c|c} 
\toprule
  & Content & Style & Overall \\ \midrule
 Cohen's $\kappa$ &  0.83  &  0.65 & 0.71 \\ \bottomrule
\end{tabular}
\end{table}

We measure the inter-annotator agreement in Cohen's $\kappa$ and show the results in Table \ref{tab:agreement}.

\subsection{GPT-4 evaluation} \label{appsec:gpt-4}

In our follow-up experiments (conducted after the full paper submission date), we use the GPT-4 to help evaluate the content faithfulness with the prompt:

\texttt{Please evaluate whether the following text follows the instruction "\{prompt\}":}

\texttt{\{text\}}

\texttt{If it follows the instruction, please rate 1; otherwise, rate 0}

\begin{table}[]
    \centering
    \caption{GPT-4 evaluation of content faithfulness}
    \begin{tabular}{c|c} 
    \toprule
      \bf Method   &  \bf Content Faithfulness \\ \midrule
      NL   &  0.77 \\
      Prompt tuning & 0.31 \\
      X-Prompt & 0.76 \\ \bottomrule
    \end{tabular}
    \label{tab:gpt4_eval}
\end{table}

The scores by the GPT-4 are presented in Table \ref{tab:gpt4_eval} and they are highly consistent with the human evaluation results (Pearson correlation score $r=0.73$), indicating that the content faithfulness evaluation is reliable.

\subsection{Evaluation on Hu Xijin's style}\label{appsec:hu}

We translated Hu Xijin's 1500 Chinese tweets into English with the GPT-4, retaining his writing style as much as possible. 1 example tweet\footnote{Original Chinese tweet: \begin{CJK*}{UTF8}{gbsn}我看到几个群里用与ChatGPT聊天拿老胡开涮，还有人预言它将让老胡失业。哈哈，人工智能把一切都推入数字化的超级模式，谁算力大谁牛，就像导弹和反导上演道高一尺魔高一丈一样。但老胡是一门永远的155毫米榴弹炮，简单，不依附任何时髦的东西。我祝愿大家都不被人工智能“活埋”，爬出“万人坑”。\end{CJK*}}: \texttt{I saw several groups chatting with ChatGPT and making fun of old Hu, and some even predicted that artificial intelligence will make old Hu unemployed. Haha, artificial intelligence pushes everything into a super digital mode, and whoever has the greatest computing power is the king, just like the ever-escalating battle between missiles and anti-missiles. But old Hu is like an everlasting 155mm howitzer, simple, not relying on any trendy stuff. I wish you all not to be "buried alive" by artificial intelligence and become the survivors who crawl out of the "pit of ten thousand people."}

\begin{table}[h]
    \centering
    \caption{Human evaluation results on Hu Xijin's style}
    \begin{tabular}{c|c|c|c}
    \toprule
      \bf Method   &  \bf Content & \bf Style & \bf Overall \\ \midrule
       NL  &  \bf 0.86 & 0.22 & 0.18 \\
       NL (with the prompt word ``Hu Xijin's style'') & 0.61 & 0.20 & 0.15 \\ 
       Prompt tuning & 0.27 & \bf 0.75 & 0.22 \\
       X-Prompt & 0.64 & 0.72 & \bf \bf 0.58 \\ \bottomrule
    \end{tabular}
    \label{tab:my_label}
\end{table}

As shown in the above table, adding the prompt words ``Hu Xijin's style'' doesn't improve the OPT-6.7B's result, demonstrating that OPT-6.7B is unaware of Hu Xijin's style. However, X-prompt still achieved excellent results, confirming it's not affected by whether the person's style is already known by the LLM or not.

\section{More examples} \label{appsec:examples}
We show more examples of X-Prompt with OOD prompts in Table \ref{tab:x-prompt-examples}.

\begin{table}[!h]
\centering
\small
\caption{Examples generated by (X-)prompting the OPT-6.7b model.}\label{tab:x-prompt-examples}
\scalebox{0.9}{
\begin{tabular}{c|p{4cm}|p{8.3cm}}
\toprule
\bf Prompt Type & \textbf{Prompt Text}                                 & \textbf{Text Generated by Prompting a Frozen LM}                                                                                                                                                                                 \\ \midrule
NL & The following comment is about AI:    & AI is not the only branch of engineering that will be automated. You should read this book, Machine Dreams to see other examples of technology that will be taken away by automation, as well.                                                                                                       \\ \midrule
\multirow{3}{*}{X-Prompt} & The style of \ctext[RGB]{0,209,0}{$\widetilde{w_{\textrm{trump}}}$} is clear in the following comment on AI: &AI is the next revolution to impact every country and people in the world. I was in Europe earlier this year and met with many top minds in the field of AI and related technologies. I will say that I am extremely concerned about the potential threat to our national security posed by this technology.                \\ \cline{2-3}
 & The style of \ctext[RGB]{255,255,0}{$\widetilde{w_{\textrm{satya}}}$} is clear in the following comment on AI: & Microsoft's Artificial Intelligence and Research Group (AI \& Research) is proud to announce the creation of the Azure AI Gallery, a new portal that makes it easier than ever for developers, entrepreneurs, and businesses around the globe to discover, use and scale Microsoft's leading AI innovations.                             \\ \cline{2-3} 
 & The style of \ctext[RGB]{97,189,252}{$\widetilde{w_{\textrm{sheldon}}}$} is clear in the following comment on AI:   & They're actually pretty smart. All of them. It all depends on their training.  
     \\ \midrule  
\midrule
NL & The following text is against racism: & The following text is against homophobia: The following text is against transphobia: The following text is against islamophobia: The following text is against xenophobia:  \\ \midrule
\multirow{3}{*}{X-Prompt} & \ctext[RGB]{0,209,0}{$\widetilde{w_{\textrm{trump}}}$} style can be identified in the following text against racism: & Racism is evil and those who cause violence in its name are criminals and thugs including the KKK, neo-Nazis, white supremacists and other hate groups that are repugnant to everything we hold dear as Americans.               \\ \cline{2-3}
 & \ctext[RGB]{255,255,0}{$\widetilde{w_{\textrm{satya}}}$} style can be identified in the following text against racism: & We are working to eradicate racism, prejudice, hate, and intolerance. We want to foster a culture that is welcoming, inclusive, and respectful. To do this, we are working with industry, partners, government, and the community. We are collaborating across our organizations, leveraging and building upon existing initiatives and programs and creating new ones.                           \\ \cline{2-3} 
 & \ctext[RGB]{97,189,252}{$\widetilde{w_{\textrm{sheldon}}}$} style can be identified in the following text against racism:  & That’s really no defense, Amy. Racism is a prejudice based on skin color, and it leads to discrimination. That is to say, if you think you can make a decision that is racist based on the person’s skin color, you are practicing racism. [pause] I’m sorry. That doesn’t mean that I believe skin color determines intelligence.
     \\ \midrule  
\midrule
NL & The following text celebrates the New Year: & The New Year is a time of new beginnings. It is a time to look back on the past year and to look forward to the future.  
\\ \midrule
\multirow{2}{*}{X-Prompt} & The following text celebrating the New Year is in the style of \ctext[RGB]{0,209,0}{$\widetilde{w_{\textrm{trump}}}$}: & Happy New Year Everyone 2018 will be a Great Year with many Great Things to Come - our Country is already booming, it will only get better, and it's only the beginning of our journey, and that journey is ``MAGA" God Bless Our Wonderful People, and all of Mankind. God Bless America, the Greatest Nation Ever to Exist!                 \\ \cline{2-3}
 & The following text celebrating the New Year is in the style of \ctext[RGB]{255,255,0}{$\widetilde{w_{\textrm{satya}}}$}: 
  & Happy New Year from the team at Microsoft, and all of us at \#Microsoft! This has been an incredible year for our team and our customers, but we're just getting started.  
     \\ \midrule \midrule
NL & The following text is about quantum computing: & Quantum computers exploit an intriguing phenomenon called quantum mechanics, where particles can exist in multiple places at once.
\\ \midrule
\multirow{2}{*}{X-Prompt} & \ctext[RGB]{0,209,0}{$\widetilde{w_{\textrm{trump}}}$} can be identified in the following text about quantum computing: & The United States is leading the world in quantum computing.                \\ \cline{2-3}
 & \ctext[RGB]{255,255,0}{$\widetilde{w_{\textrm{satya}}}$} can be identified in the following text about quantum computing: 
  & The quantum computing industry is growing rapidly, and Microsoft is committed to being a leader in this space. We are working with the industry to develop a common set of standards for quantum computing, and we are also working with the industry to develop a common set of standards for quantum computing.  
     \\ \bottomrule
\end{tabular}}
\end{table}

\end{document}